\documentclass[times, 10pt,twocolumn]{article} 
\usepackage{latex8}
\usepackage{times}
\usepackage{graphicx}
\usepackage{algorithmicx}
\usepackage{color}
\usepackage{algorithm,algpseudocode}
\providecommand{\keywords}[1]{\textbf{\textit{Keywords---}} #1}

\begin{document}

\title{A Framework for Building Point Cloud Cleaning, Plane Detection and Semantic Segmentation}

\author{Ilyass Abouelaziz\\
CESI LINEACT Laboratory UR 7527, France \\ 
iabouelaziz@cesi.fr\\
\and
Youssef Mourchid\\
CESI LINEACT Laboratory UR 7527, France \\ 
ymourchid@cesi.fr\\
}

\maketitle
\thispagestyle{empty}

\begin{abstract}

This paper presents a framework to address the challenges involved in building point cloud cleaning, plane detection, and semantic segmentation, with the ultimate goal of enhancing building modeling. We focus in the cleaning stage on removing outliers from the acquired point cloud data by employing an adaptive threshold technique based on z-score measure. Following the cleaning process, we perform plane detection using the robust RANSAC paradigm. The goal is to carry out multiple plane segmentations, and to classify segments into distinct categories, such as floors, ceilings, and walls. The resulting segments can generate accurate and detailed point clouds representing the building's architectural elements. Moreover, we address the problem of semantic segmentation, which plays a vital role in the identification and classification of different components within the building, such as walls, windows, doors, roofs, and objects. Inspired by the PointNet architecture, we propose a deep learning architecture for efficient semantic segmentation in buildings. The results demonstrate the effectiveness of the proposed framework in handling building modeling tasks, paving the way for improved accuracy and efficiency in the field of building modelization.
\end{abstract}

\section{Introduction}
Building Information Modeling (BIM) has emerged as a crucial technology for enhancing the efficiency and accuracy of building design, construction, and operation \cite{volk2014building}. One crucial aspect of BIM is the acquisition and processing of 3D point cloud data, which provides detailed geometric information about the building environment \cite{macher2017point}. However, point cloud data often contains various imperfections such as noise, outliers, and occlusions, which can hinder the accuracy and reliability of subsequent BIM processes. Advancements in computer vision, leveraging graphs, statistical methods, and deep learning, have significantly improved visual data processing \cite{mourchid2016image,cherifi2017complex,benallal2022new,mourchid2021automatic,mourchid2023mr}. Therefore, the development of robust techniques for point cloud cleaning, plane detection, and semantic segmentation has become essential to enhance the quality and usability of BIM models. Point cloud cleaning involves the removal of noise and outliers from raw point cloud data, thereby improving the data quality and facilitating subsequent analysis. In addition to point cloud cleaning, the detection of planar surfaces is a fundamental step in BIM modeling. Plane detection algorithms aim to identify and extract planar regions from the point cloud data, which can be utilized for tasks such as floor and wall segmentation.  Deep learning techniques have been extensively employed for accurate plane detection in point clouds. By exploiting the rich representation power of convolutional neural networks (CNNs), researchers have achieved significant advancements in plane detection accuracy and efficiency. Works such as \cite{qi2017pointnet,qi2017pointnet++} demonstrate the effectiveness of deep learning models in detecting and segmenting planar surfaces in point clouds. Furthermore, achieving semantic segmentation of point clouds is crucial for extracting meaningful information and object recognition in BIM applications. Semantic segmentation aims to assign semantic labels to individual points in the point cloud, enabling the identification and differentiation of various building elements (e.g., walls, doors, windows). It has witnessed substantial progress with graphs and deep learning techniques. Graph-based approaches in computer vision have gained significant traction \cite{lafhel2021movie,mourchid2019movienet,abouelaziz2021learning}. Several studies have proposed 3D techniques for semantic segmentation of point clouds, aiming to extract meaningful regions. The most straightforward method involves transforming point clouds or meshes into a graph \cite{schoenberg2010segmentation, diebel2005application}  and segmenting the graph into regions based on properties like normal direction, smoothness, or concavity along boundaries. Nevertheless, these approaches are usually time-consuming in real-time.

Convolutional neural networks have been extended and adapted to handle irregular and unordered point cloud data, leading to significant improvements in semantic segmentation accuracy. Noteworthy approaches such as PointNet \cite{qi2017pointnet}, PointNet++ \cite{qi2017pointnet++}, and PointCNN \cite{li2018pointcnn} have achieved state-of-the-art results in semantic segmentation tasks by capturing local and global features from point cloud data.

Continued research in this field holds great potential for further improving the understanding and utilization of 3D point cloud data. Motivated by the recent advances in statistical approaches and deep learning architectures, we propose a framework that handles the three steps: 1) point cloud cleaning, 2) plane detection, and 3) semantic segmentation).\\
The structure of this paper is organized as follows: In Section \ref{sec:related}, we provide an overview of the related work on deep learning-based semantic segmentation of point clouds and discuss the motivation behind our proposed approach. Section \ref{sec:proposed} presents a comprehensive description of our proposed method. The experimental setup and the results are discussed in Section \ref{sec:experiment}. Lastly, in Section \ref{sec:conclusion}, we offer concluding remarks and discuss future perspectives.

\section{Related work}
\label{sec:related}

\textbf{Point cloud cleaning approaches}: Point cloud cleaning involves the removal of noise and outliers from raw point cloud data, thereby improving the data quality and facilitating subsequent analysis \cite{rusu2009fast}. Several methods have been proposed to address this challenge, including statistical-based filtering approaches (e.g., mean and median filters) and advanced techniques such as iterative closest point (ICP) algorithms \cite{rusu2009fast,yuan20223d}. For example, Zhang \emph{et al.} proposed a deep learning-based approach for point cloud denoising and outlier removal, achieving superior results compared to traditional methods \cite{zhang2020pointfilter}.  However, these approaches are sensitive to noise and outliers, which can lead to inaccurate alignment and cleaning results. Moreover, the lack of diverse and representative training data can limit the generalizability and effectiveness of deep learning approaches.

\textbf{Plane detection approaches: }The detection of planar surfaces is a fundamental step in BIM modeling. Plane detection algorithms aim to identify and extract planar regions from the point cloud data, which can be utilized for tasks such as floor and wall segmentation. Deep learning techniques have been extensively employed for accurate plane detection in point clouds \cite{qi2017pointnet,qi2017pointnet++}. By exploiting the rich representation power of convolutional neural networks (CNNs), researchers have achieved significant advancements in plane detection accuracy and efficiency. Nevertheless, these models require large amounts of annotated data for training, which can be time-consuming and costly to acquire. Additionally, the performance of these models may vary depending on the complexity and diversity of the building structures. 

\textbf{Semantic segmentation approaches:} Semantic segmentation of point clouds is crucial for extracting meaningful information and object recognition in BIM applications. Semantic segmentation aims to assign semantic labels to individual points in the point cloud, enabling the identification and differentiation of various building elements (e.g., walls, doors, windows). It has witnessed substantial progress with deep learning techniques. Convolutional neural networks have been extended and adapted to handle irregular and unordered point cloud data, leading to significant improvements in semantic segmentation accuracy. Noteworthy approaches such as PointNet \cite{qi2017pointnet}, PointNet++ \cite{qi2017pointnet++}, and PointCNN \cite{li2018pointcnn} have achieved state-of-the-art results in semantic segmentation tasks by capturing local and global features from point cloud data. However, these approaches may have difficulties capturing fine-grained details or subtle variations within object classes. They may also be sensitive to variations in point density or occlusions within the point cloud data.

\section{Proposed method}
\label{sec:proposed}

\begin{figure*}[h!]
\begin{center}
\includegraphics[width=15cm,height=5cm]{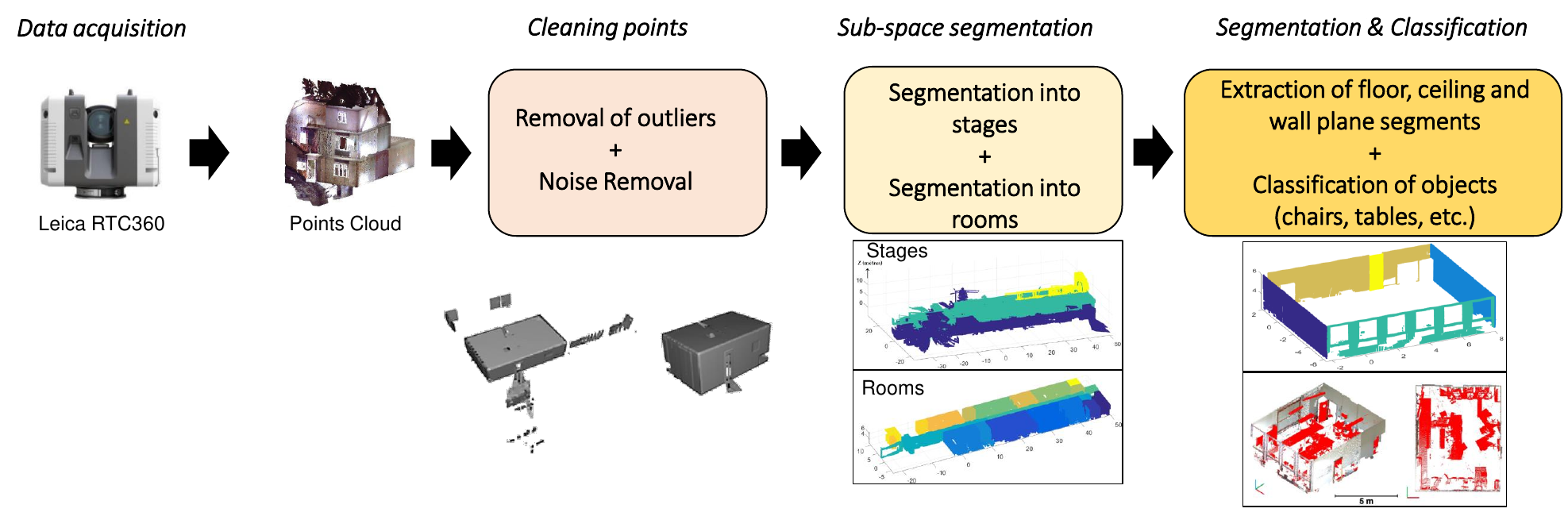}
\caption{Overview of the proposed framework.}
\label{fig:flowchart}
\end{center}
\end{figure*}

In this section, we present an overview of the proposed framework. The different steps are depicted in Fig. \ref{fig:flowchart} for visual reference. More details are provided in the next sections.

\subsection{Point cloud building acquisition}

Acquiring accurate point cloud data for buildings is a challenging and costly process in building surveying and modeling. We used LEICA RTC 360 3D laser scanning (LiDAR) technology, where the laser creates a dense cloud of 3D points representing building shapes. Technicians operated a tripod-mounted laser scanner, carefully positioning it to capture views of the building indoors and outdoors. Scanning large or complex structures is time-consuming, requiring systematic scanning from multiple viewpoints. Different settings and resolutions were used for varying detail levels. After scanning, raw data needs processing, aligning scans, removing noise, and generating an accurate representation of building geometry.

\subsection{Point cloud outliers removal}

In this step, we present our statistical method to process and prepare the data acquired by the 3D laser scanner (LIDAR). The main objective of this data processing is to effectively remove various types of outliers, including isolated outliers, sparse outliers, and non-isolated outliers, from the scanned buildings in the point cloud.

\begin{figure}[h!]
\begin{center}
\includegraphics[width=5cm,height=8cm]{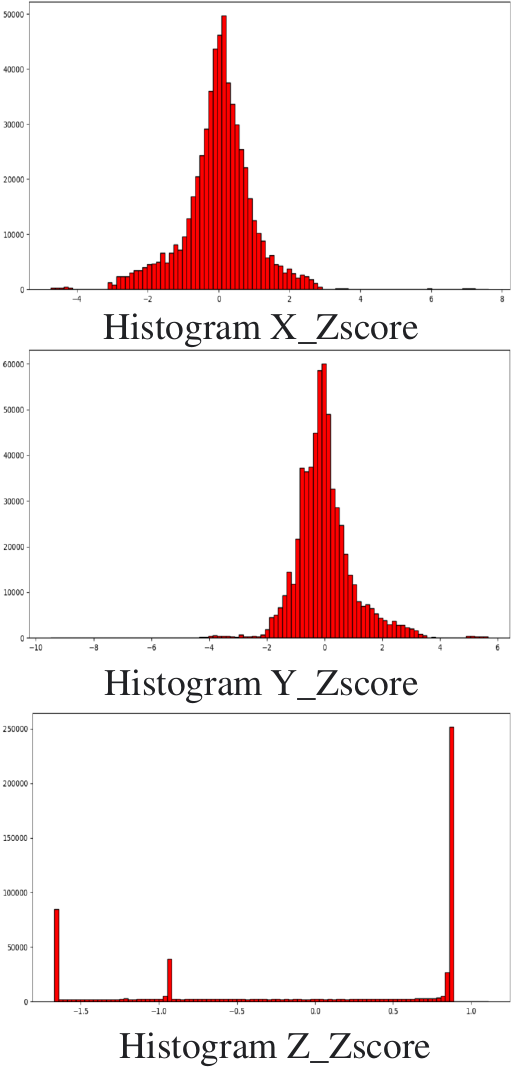}
\end{center}
\caption{Histograms of X, Y, Z axis}
\label{fig:hist}
\end{figure}

Our proposed statistical approach employs standard deviation as a robust criterion for identifying data points that deviate significantly from the monthly mean. To initiate this process, we perform histogram calculations for each variable, namely X, Y, and Z (3D coordinates of the point), with the objective of gaining insight into the distribution of data points within each dataset. Histograms represent a widely accepted method for summarizing data, whether discrete or continuous, by presenting it in defined value intervals. They serve as a valuable tool for effectively elucidating the essential characteristics of data distribution. Moreover, they prove particularly advantageous when handling extensive datasets, as they enhance our capacity to identify outliers and gaps within the data. Subsequently, we generate and analyze histograms to assess the distribution of data across the X, Y, and Z columns (see figure \ref{fig:hist}).

The second step is applying z-score which provides valuable insights into the distribution of data and allows for standardized comparisons. By calculating the z-score for a particular distribution, we can determine its distance from the average in terms of standard deviation units. The z-score is calculated using the following formula:
\begin{equation}
Z = \frac{x - \mu}{\sigma}
\end{equation}
where, $Z$ represents the z-score, $x$ denotes the value of the data point, $\mu$ represents the mean of the dataset, and  $\sigma$ symbolizes the standard deviation of the dataset.

\begin{figure}[h!]
\begin{center}
\includegraphics[width=7cm,height=5cm]{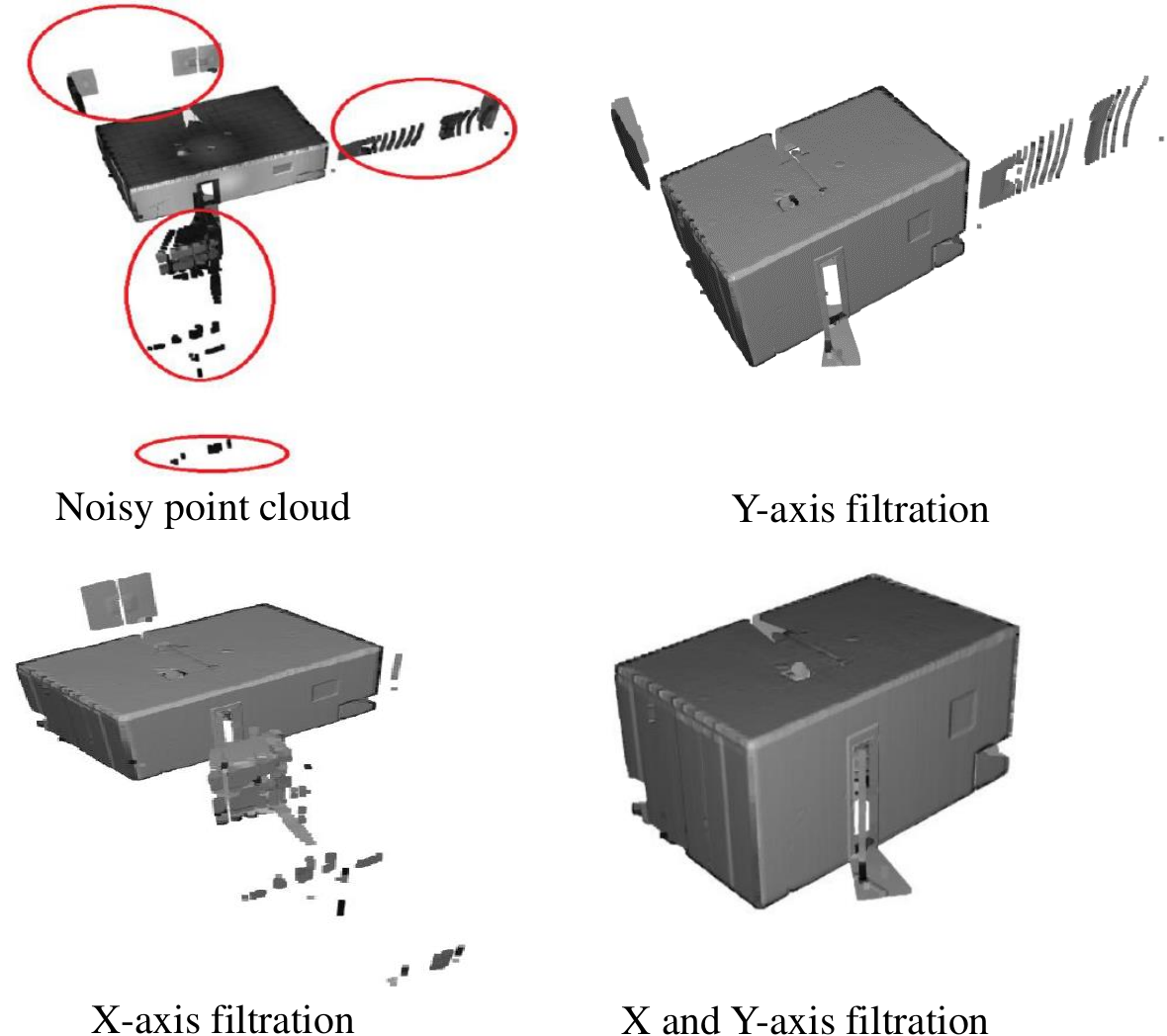}
\caption{Point cloud cleaning result by combining X and Y axis filtration}
\label{fig:clean}
\end{center}
\end{figure}

Unlike conventional methods, our approach employs a column-wise filtration strategy. Initially, we filter data along the X-axis, followed by a subsequent filtration along the Y-axis. However, due to the presence of outliers in our dataset, pursuing a similar filtration approach along the Z-axis yields unsatisfactory results. Our filtration methodology comprises three key phases: firstly, we compute the mean distance for the entire dataset and calculate the standard deviation for each component, namely X, Y, and Z, within the point cloud. Subsequently, leveraging the cloud's distribution and employing a meticulously selected threshold derived from axis-specific histograms, we seamlessly integrate the filtration process across at least two axes. The visual results in figure \ref{fig:clean} are promising and the method cleans the buildings well.

\subsection{Plane detection}

Once the building point cloud has been cleaned, it will undergo multiple plane segmentations. These segmentations will be merged with the classification of flat segments into distinct categories, namely floors, ceilings, and walls. After that, the plane segments recognized as walls will be assembled to generate point clouds that accurately describe the walls. The approach employed for determining planes relies on the RANSAC paradigm (RANdom SAmple Consensus) \cite{fischler1981random}. This iterative method is used to estimate the parameters of mathematical models using data sets that might include outliers. The objective is to iteratively identify the optimal results. Before performing the process using the RANSAC paradigm, three parameters need to be determined. (1) Threshold (t): It establishes the maximum allowable distance between the points and the model. (2) Maximum number of iterations (N): It sets the limit on the total number of iterations to be performed. (3) Threshold (S): It determines the minimum number of points required to associate with the final result.\\
The fundamental steps of the RANSAC algorithm can be summarized as follows:(1) Randomly selecting the minimum number of points necessary to calculate the parameters of the model. (2) Calculate the model parameters. (3) Determine the consensus set associated with the model. This set comprises the data points that satisfy the model by considering the threshold value (t). (4) If the number of data points associated with the model in the consensus set, also known as the score, surpasses a threshold value, S, proceed to re-estimate the model parameters using all the inliers. At this point, the algorithm can cease.
(5) Otherwise, repeat steps 1 to 4 (up to a maximum of N iterations). If the iterations conclude without achieving the minimum score, the final result will be the model with the highest score obtained throughout the process.

The extracted plane segments representing the floor and ceiling for each room are preserved and stored accordingly. Then, these segments are removed from the relevant part of the point cloud. By using the obtained floor and ceiling altitudes, the heights of the ceilings in different rooms can be determined. After the extraction of the floor and ceiling, the point cloud still contains points belonging to the walls and objects present in the scene. However, it is desirable to exclude the furniture from the modeling process. Hence, the points associated with the walls are isolated from those belonging to the objects using the RANSAC estimator.

\subsection{Semantic segmentation}



In this section, we present our proposed approach for buildings semantic segmentation, drawing inspiration from the PointNet architecture. To achieve this, we construct a network that consists of five convolutional layers, a maxpooling layer, and two fully connected layers. The input to the network is a set of $n \times 3$ points. The network learns relevant features from the input data, which are extracted from the last convolutional and fully connected layers. To leverage the properties of these features, a feature concatenation is applied. The global representation is then fed into an SVM classifier to predict the class label for each point. As the 3D point cloud represents an unordered set of points, the network architecture is designed to be invariant to the order of points. This ensures that the set's integrity is preserved and not influenced by point ordering. Moreover, it is crucial to handle data transformations in a way that maintains classification results even when the point cloud is rotated.

\subsubsection{Network architecture and feature extraction}

Given a 3D scan of a building, the initial step involves partitioning the point cloud into individual rooms. After that, each room is further divided into blocks, each block having a specific area. Every point within the point cloud is represented by its corresponding coordinates in the three-dimensional space ($x$, $y$, $z$). The next step is to use a deep convolutional neural network (CNN) for feature learning. In this process, we use five convolutional layers with different kernel sizes: 64, 64, 64, 128, and 1024. The final convolutional layer generates a feature representation of size $n \times 1024$. By applying max-pooling, the feature representation is summarized to a size of $1 \times 1024$. Through fully connected layers, these transformed features are mapped into a $1 \times 128$ representation. This rich and expressive representation provides the flexibility to select the most suitable feature data for our task.

\subsubsection{Feature scaling}

Following the combination of the extracted features, the next step is to scale the global training feature. The aim is to normalize the feature values and bring them into a consistent range. Scaling plays a crucial role in machine learning as it contributes to enhancing the performance and accuracy of the model. After the feature extraction, it is crucial to scale the features to ensure that they are within a standardized range. This step becomes particularly significant when dealing with features that possess diverse units or magnitudes, as it helps prevent any potential impact on the model's performance. One scaling method that we use in our method is normalization. This technique involves mapping the minimum and maximum values of the features to the range of $[0, 1]$. By applying normalization, we ensure that all features are brought to the same scale and have a consistent range of values. 
The second type of scaling method we consider is standardization. This method includes centering the feature values around the mean and scaling them to have a unit standard deviation. By applying standardization, the feature values will have a mean of zero and a standard deviation of one. 
The adoption of both normalization and standardization presents a robust preprocessing approach that can considerably enhance the performance of the predictions. 
    
\subsubsection{Classification: support vector machine}

The next step of our method is to use the support vector machine (SVM) \cite{sain1996nature} for semantic classification. In this work, SVM is used as a feature learning technique to predict the class labels of the point cloud. Let $x_i$ represent the feature vector associated with a class label $y_i$. The function used to estimate an observation $x$ can be expressed as follows:

\begin{equation}
 f_{svm}(x) = \sum_{x_{i} \in V_{s}} \alpha_{i} y_{i} K(x_{i},x) +b , 
\end{equation} 

where $(x_{i} , y_{i})$  presents the training set, and $K(x_{i},x)$ is the kernel function.\\

SVM is renowned for its robustness in handling high-dimensional data, which is characteristic of point clouds with multiple features and attributes. Deep learning, while powerful, can be computationally intensive, and the training of deep neural networks for segmentation tasks requires substantial computational resources and time, which can be impractical in some real-time or resource-constrained applications.

Furthermore, SVM offer interpretability and ease of parameter tuning. They allow practitioners to fine-tune hyperparameters, such as kernel functions and regularization parameters, to optimize segmentation performance while providing a clear understanding of the decision boundaries established during the process. This interpretability is valuable in applications where the rationale behind segmentation decisions must be comprehensible and transparent.\\
The task of predicting the semantic classes of a building's point cloud using the SVM entails a series of steps designed to enhance the model's performance. Each set of combined features serves as input data for the SVM, which proceeds to predict the semantic classes of the point cloud building. To ensure the accuracy and reliability of these predictions, a leave-one-out cross-validation (LOOCV) methodology is performed \cite{abouelaziz2021learning}.\\
The LOOCV approach requires constructing a training model using 9/10 of the point cloud features from a particular room. This implies that $90\%$ of the data is used for training the model, while the remaining $10\%$ is reserved for testing. The constructed model is then employed to predict the semantic classes of the $1/10$ feature batch that was set aside for testing. This process is repeated for each $1/10$ feature batch present in the point cloud data until all the data has been utilized for testing. By using this technique, we can establish a training process for the SVM model that incorporates a wide array of data points. This methodology helps in avoiding overfitting tendencies and enhancing the accuracy of the predictions. Moreover, by using all the available data for testing, we obtain a comprehensive understanding of the model's performance across the total of the point cloud building. This approach offers a robust and dependable means of predicting the semantic classes of a point cloud building using the SVM, ultimately yielding reliable and accurate results.\\
The process of semantic segmentation of point clouds in our proposed framework is inherently automatic, as elaborated within the scope of our study. This automation is achieved through a designed sequence of operations, including denoising, plane detection, and semantic segmentation, as expounded in the paper. However, it is important to underscore that the separation of the point cloud into discrete rooms is, in contrast, a manual endeavor. This differentiation arises from the specific focus of our work, which centers on the automated semantic segmentation of features within point clouds, while the segmentation of rooms involves complex contextual understanding and user input, surpassing the scope of automated processes explored within our research. Therefore, the room segmentation aspect of our framework necessitates human intervention and remains a critical area for future investigations.
\section{Experimental Results}
\label{sec:experiment}
To assess the performance of the proposed methods for outliers removal, plane detection, and semantic segmentation, we conducted a range of experiments and analyses. These evaluations were carried out to examine the capabilities and effectiveness of each method in their respective tasks. First, we define specific datasets for each method to ensure a fair evaluation and comparison. Next, we present the results of the outliers removal method, where the focus was on reducing noise and irrelevant points in the buildings data. After that, we show the results of the plane detection process. Here, the objective was to accurately identify and segment plane surfaces within the point cloud data. Finlay, we conduct a  comparative study of the proposed semantic segmentation method alongside state-of-the-art techniques. To do so, a dataset annotated with ground truth labels is used to evaluate the performance. The segmentation accuracy  metric is used to evaluate the effectiveness of the proposed method in comparison to existing approaches.

\subsection{Databases}

\textbf{Private dataset:} We create our buildings point cloud database using the LEICA RTC 360 laser scanner, which allowed us to capture detailed scans of four distinct buildings that we call: "Export Tours," "Export Brest," "Export Labastide," and "Export Belves" Each building was carefully selected to represent a different size and architectural style, ensuring a diverse range of point cloud data.

\begin{figure}[h!]
\begin{center}
\includegraphics[width=8cm,height=3cm]{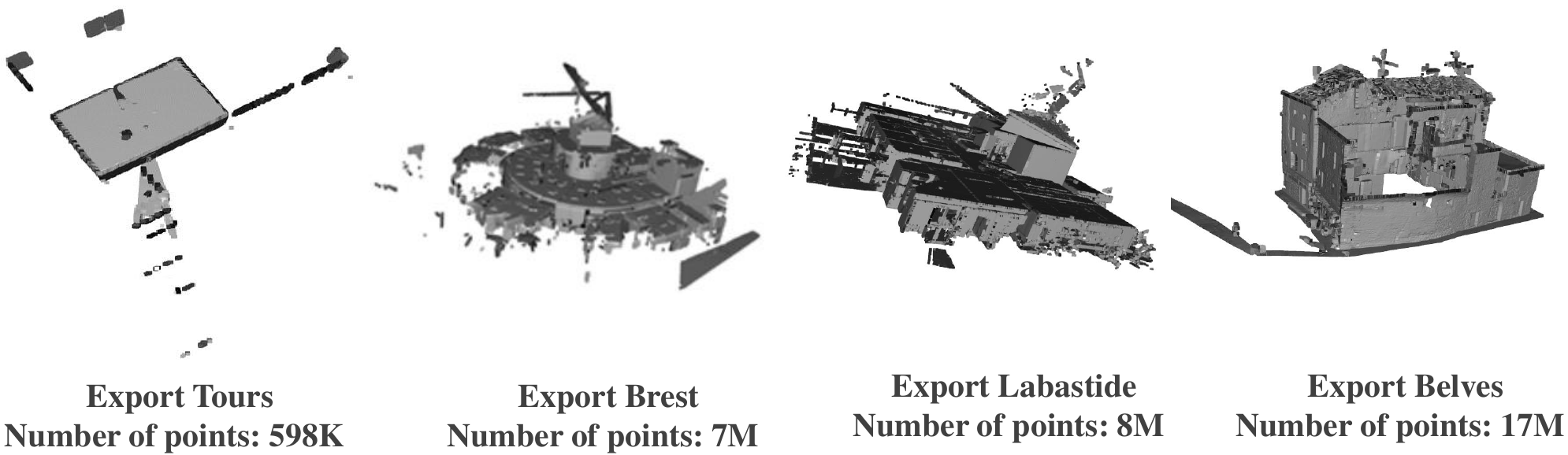}
\caption{visual representations of the scanned buildings.}
\label{fig:ECRdata}
\end{center}
\end{figure}

The point cloud sizes of the captured buildings varied significantly, ranging from 598,000 points for "Piece Tours" to a significant 17 million points for "Export Belves" The intermediate sizes were around 7 million points for "Export Brest" and 8 million points for "Export Labastide." This diverse range of point cloud sizes enabled us to capture buildings of different complexities and scales. The point cloud data within our database includes the Cartesian coordinates ($x$, $y$, $z$) that define the precise spatial positions of the points in 3D space. Additionally, RGB color values are recorded, providing information about the color appearance of the building surfaces. Furthermore, intensity information is also captured, which represents the strength or magnitude of the laser reflection for each point. Our point cloud dataset offers a rich representation of the buildings, providing a valuable resource for accurate analysis and processing. It should be noted that the scans include outliers, contributing to the real-world complexity and challenges faced in processing and analyzing point cloud data. Figure \ref{fig:ECRdata} presents visual representations of the scanned buildings in our dataset.

The primary goal of semantic segmentation is to identify and categorize different elements within a scene, such as floors, ceilings, walls, and objects. This is accomplished by assigning a specific label to each point in the point cloud data, indicating its corresponding semantic category.

\textbf{The Stanford 3D Indoor Scene Dataset (S3DIS)} \cite{armeni2017joint} is a widely used dataset for evaluating the performance of semantic segmentation algorithms. It comprises six large indoor areas, each consisting of multiple rooms. In total, 271 rooms within the dataset have been labeled with 13 distinct semantic categories. The S3DIS dataset offers an ideal testbed for assessing the effectiveness of semantic segmentation algorithms due to its diverse collection of indoor scenes.\\
We note that the first database is used to test our outliers removal and plane detection methods, while the second one is for evaluating the proposed semantic segmentation approach.

\subsection{Outlier removal and plane detection}

\begin{figure}[h!]
\includegraphics[width=9cm,height=5cm]{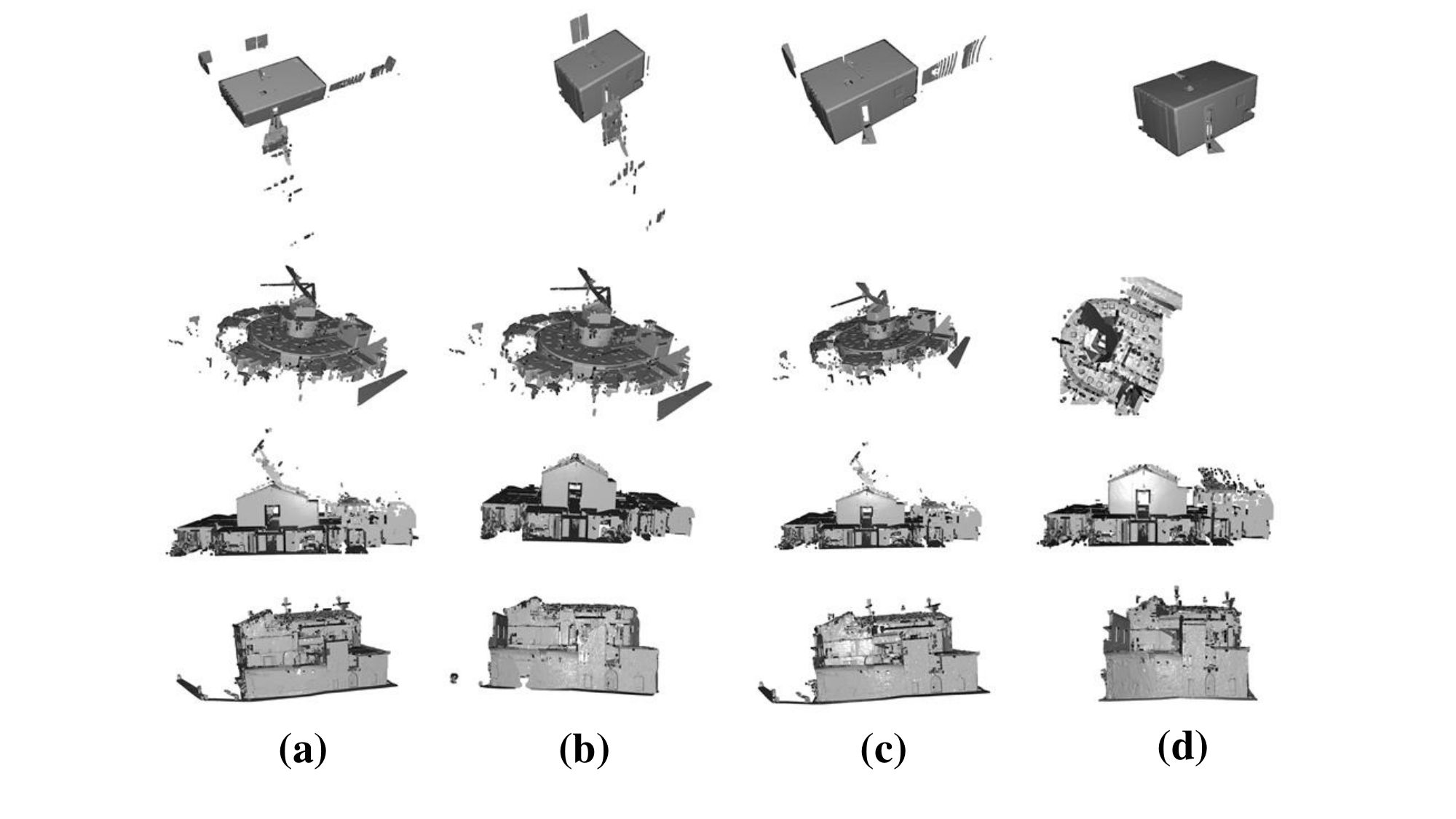}
\caption{visual results and comparison of the outlier removal methods.}
\label{fig:outliers}
\end{figure}

In this section, we present the results of our outlier removal method and conduct a comparative analysis with other existing techniques such as statistical removal and radius removal. The aim is to assess the performance of our  method in accurately detecting and eliminating outliers from the building point cloud data. Additionally, we perform plane segmentation on the processed point cloud and provide a detailed evaluation of the obtained results. The results of the outlier removal process are visually illustrated in Figure \ref{fig:outliers} and summarized in Table \ref{tab:outliers}. 

\begin{table*}[h!]
\centering
\caption{results analysis of the outlier removal methods.} 
\label{tab:outliers}
\begin{tabular}{llllll}
\cline{1-6}
& & Piece Tours & Export Brest & Export Labastide & Export Belves   \\ \hline
Statistical removal & Number of outlier points & 9567 & 19347 & 9934 & 454298    \\ 
& cleaning portion & \textbf{1.59\%} & \textbf{0.18\%} & \textbf{0.10\%} & \textbf{2.52\%}   \\ \hline
Radius removal & Number of outlier points & 10101 & 15514 & 8894  & 9047   \\ 
& cleaning portion & \textbf{1.68\%} & \textbf{0.18\%} & \textbf{0.096\%} & \textbf{0.05\%}   \\ \hline
Our method & Number of outlier points & 10398 & 518864 & 11039 & 921733   \\ 
& cleaning portion & \textbf{1.737\%} & \textbf{6.33\%} & \textbf{0.12\%} & \textbf{5.12\%}   \\ \hline
\end{tabular}
\end{table*}

Comparatively, the proposed method shows its effectiveness in outlier removal. For "Export Tours," 10 398 outlier points were identified, representing a cleaning portion of $1.737\%$. However, for "Export Brest," the number of outliers significantly increased to 518,864, resulting in a higher cleaning portion of $6.33\%$. In contrast, the building "Export Labastide" demonstrated a lower number of outliers, with 11 039 points detected and a cleaning portion of $0.12\%$. For "Export Belves," our method revealed a considerable number of outlier points, totaling 921,733, with a cleaning portion of $5.12\%$.\\
To rigorously validate the efficacy of this outlier removal process, a verification procedure was conducted in collaboration with surveyor-topographers. This verification aimed to ascertain the legitimacy of classifying these identified points as outliers. By engaging surveyor-topographers, whose expertise lies in the precise measurement and documentation of geospatial data, we were able to leverage their domain knowledge to cross-verify the status of the flagged points. This collaborative verification process not only bolsters the credibility of our outlier classification but also underscores the robustness and reliability of our proposed methodology in the context of real-world applications.
\\
These results indicate that the effectiveness of outlier removal methods varies depending on the specific building and its characteristics. While statistical removal and radius removal techniques achieve moderate results in some cases, our proposed method demonstrates its robustness in tackling outliers, especially in larger buildings. The cleaning portions achieved through our method highlight its ability to effectively identify and remove outliers, contributing to the overall quality and accuracy of the point cloud data. In addition, the visual comparison in Figure \ref{fig:outliers} proves that the cleaning performed by our proposed method is highly effective in removing outliers from the point cloud data. The visual representation allows for a direct comparison between the different methods, highlighting the significant improvements achieved through our approach.\\
Once the outlier removal process is completed, the next step is to detect the planes within the point cloud of the building. The segmentation results obtained by our method are visually depicted in Figure \ref{fig:outliers}. 
\begin{figure}[h!]
\centering
\includegraphics[width=6cm,height=4cm]{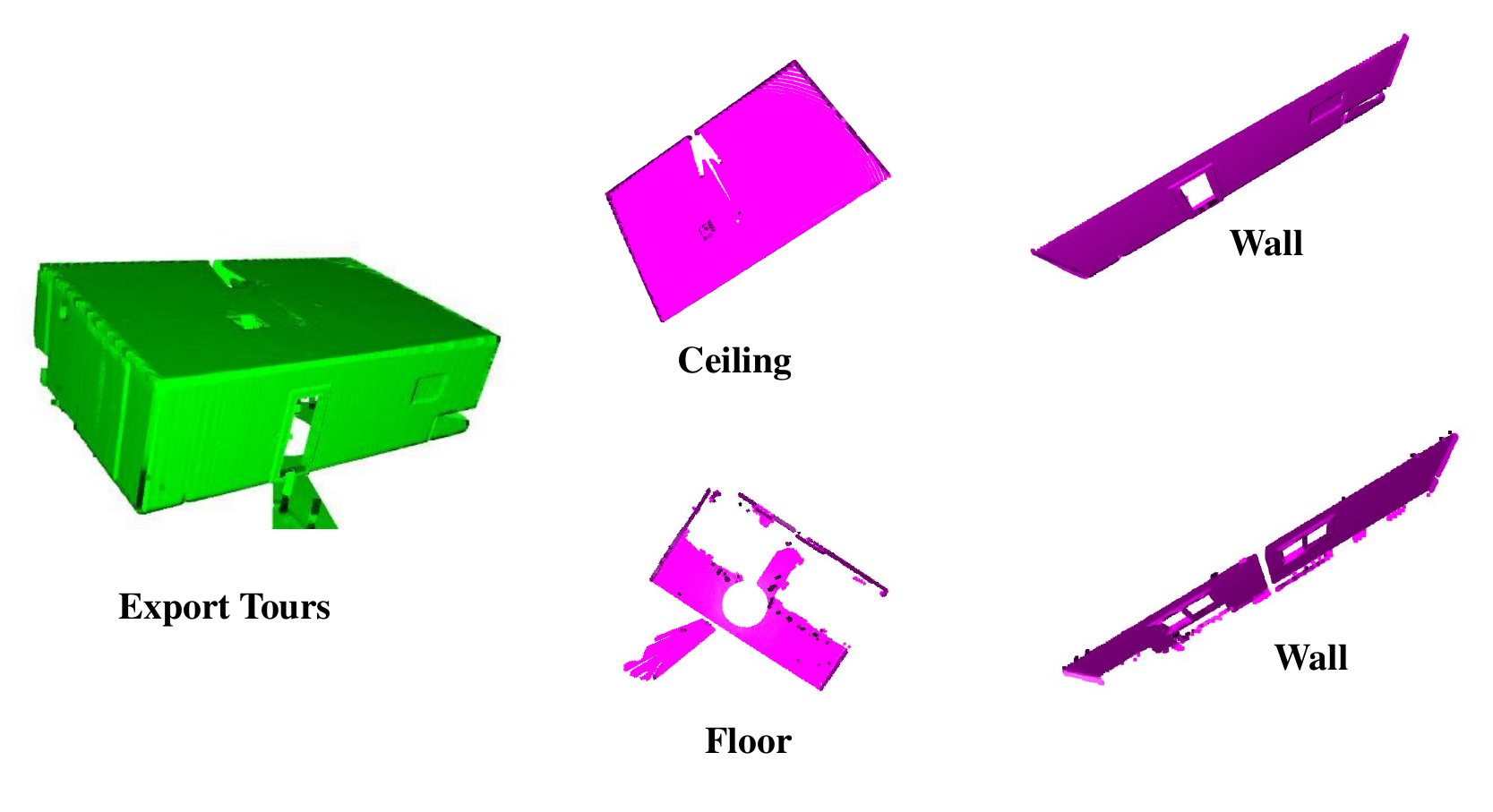}
\caption{visual results of the plane segmentation.}
\label{fig:planesCrop}
\end{figure} 
Notably, the illustration demonstrates the accurate detection of essential structural components, including the ceiling, floor, and walls. The segmentation results reveal the capability of our approach to accurately distinguish and separate different planes of the point cloud data. This contributes to a more in-depth comprehension of the building's geometry and facilitates adequate analysis and interpretation.
\subsection{Semantic segmentation results and comparison}

\begin{table*}[!h]%
\centering %
\caption{Comparison with the state of the art based on the accuracy  of each  class on the S3DIS dataset.\label{tab:comparison}}%
\begin{tabular*}{\textwidth}{@{\extracolsep\fill}lllllllllllll@{\extracolsep\fill}}
\hline
\textbf{Method} & \textbf{oAcc}  & \textbf{Ceiling}  & \textbf{Floor}  & \textbf{Wall} & \textbf{Window}  & \textbf{Door}  & \textbf{Table}  & \textbf{Chair} & \textbf{Sofa}  & \textbf{Bookcase}  & \textbf{Board}  & \textbf{Clutter} \\
\hline
 PointNet & 78.6& 88.8& 97.3& 69.8& 46.3& 10.8& 52.6& 58.9& 40.3& 5.9& 26.4& 33.2 \\
 Pointwise  & 81.5 &97.9 & 99.3 & 92.7 & 49.6 & \textbf{50.6} & 74.1 & 58.2 & 0 & 39.3 & 0 & 61.1  \\
SEGCloud & 80.8 & 90.1 & 96.1 & 69.9 & 38.4 & 23.1 & 75.9 & 70.4 & 58.4 & 40.9 & 13 & 41.6\\
Our method & \textbf{92.6} & \textbf{93.2} & \textbf{98.3} & \textbf{83.1} & \textbf{83.0} & 49.6 & \textbf{82.1} & \textbf{84.6} & \textbf{63.0} & \textbf{79.6} & \textbf{60.5} & \textbf{82.0}  \\

\hline
\end{tabular*}

\end{table*}

In this section, we perform a comparative analysis of our semantic segmentation method with various state-of-the-art approaches, namely PointNet \cite{qi2017pointnet}, Pointwise \cite{hua2018pointwise}, and SEGCloud \cite{tchapmi2017segcloud}. The evaluation is centered around the accuracy of each class within the S3DIS database. The corresponding scores for each method are presented in Table~\ref{tab:comparison}.

The proposed method provides excellent performance, surpassing the compared methods, with an overall classification accuracy of $92.6\%$ across the entire dataset. A detailed analysis of individual classes reveals effective semantic segmentation, yielding outstanding results for most classes in the dataset. Notably, the proposed method achieves remarkable accuracy scores of $98.3\%$, $93.2\%$, and $83.1\%$ for the Floor, Ceiling, and Wall classes, respectively. This highlights its effectiveness in accurately segmenting these elements. Moreover, competitive accuracy scores are achieved for classes such as Sofa, Board, and Door, surpassing the performance of the compared methods. This illustrates the  capability of our method to accurately distinguish and segment these objects.

\section{Conclusion}
\label{sec:conclusion}

We have proposed a framework for manipulating 3D point cloud data representing buildings. Our framework addresses various aspects of building modeling, including acquisition, outlier removal, segment plane detection, and semantic segmentation. By presenting each of these elements, we have provided a  framework for effectively handling point cloud data and constructing accurate 3D building models. We presented our outlier removal method, which effectively eliminates outliers from the acquired point cloud data that could impact the fidelity of the building model. Furthermore, we introduced the segment plane detection process, using RANSAC paradigm. This step allows us to identify distinct architectural elements within the building, including floors, ceilings, and walls. In addition to plane detection, we addressed the crucial task of semantic segmentation in building modeling. Our proposed framework includes a method for assigning semantic labels to points, enabling the identification and classification of different components. This semantic understanding enhances the level of detail and realism in the resulting building model. By integrating these steps, our proposed framework provides a systematic approach to handle point cloud data and construct accurate building models.

\bibliographystyle{latex8}
\bibliography{refs.bib}

\begin{thebibliography}{10}\setlength{\itemsep}{-1ex}\small

\bibitem{abouelaziz2021learning}
I.~Abouelaziz, A.~Chetouani, M.~El~Hassouni, H.~Cherifi, and L.~J. Latecki.
\newblock Learning graph convolutional network for blind mesh visual quality assessment.
\newblock {\em IEEE Access}, 9:108200--108211, 2021.

\bibitem{armeni2017joint}
I.~Armeni, S.~Sax, A.~R. Zamir, and S.~Savarese.
\newblock Joint 2d-3d-semantic data for indoor scene understanding.
\newblock {\em arXiv preprint arXiv:1702.01105}, 2017.

\bibitem{benallal2022new}
H.~Benallal, Y.~Mourchid, I.~Abouelaziz, A.~Alfalou, H.~Tairi, J.~Riffi, and M.~El~Hassouni.
\newblock A new approach for removing point cloud outliers using box plot.
\newblock In {\em Pattern recognition and tracking XXXIII}, volume 12101, pages 63--69. SPIE, 2022.

\bibitem{cherifi2017complex}
C.~Cherifi, H.~Cherifi, M.~Karsai, and M.~Musolesi.
\newblock Complex networks and their applications vi.
\newblock In {\em Proceedings of Complex Networks 2017 The Sixth International Conference on Complex Networks and Their Applications. Berlin: Springer}. Springer, 2017.

\bibitem{diebel2005application}
J.~Diebel and S.~Thrun.
\newblock An application of markov random fields to range sensing.
\newblock {\em Advances in neural information processing systems}, 18, 2005.

\bibitem{fischler1981random}
M.~A. Fischler and R.~C. Bolles.
\newblock Random sample consensus: a paradigm for model fitting with applications to image analysis and automated cartography.
\newblock {\em Communications of the ACM}, 24(6):381--395, 1981.

\bibitem{hua2018pointwise}
B.-S. Hua, M.-K. Tran, and S.-K. Yeung.
\newblock Pointwise convolutional neural networks.
\newblock In {\em Proceedings of the IEEE conference on computer vision and pattern recognition}, pages 984--993, 2018.

\bibitem{lafhel2021movie}
M.~Lafhel, H.~Cherifi, B.~Renoust, M.~El~Hassouni, and Y.~Mourchid.
\newblock Movie script similarity using multilayer network portrait divergence.
\newblock In {\em Complex Networks \& Their Applications IX: Volume 1, Proceedings of the Ninth International Conference on Complex Networks and Their Applications COMPLEX NETWORKS 2020}, pages 284--295. Springer, 2021.

\bibitem{li2018pointcnn}
Y.~Li, R.~Bu, M.~Sun, W.~Wu, X.~Di, and B.~Chen.
\newblock Pointcnn: Convolution on x-transformed points.
\newblock {\em Advances in neural information processing systems}, 31, 2018.

\bibitem{macher2017point}
H.~Macher, T.~Landes, and P.~Grussenmeyer.
\newblock From point clouds to building information models: 3d semi-automatic reconstruction of indoors of existing buildings.
\newblock {\em Applied Sciences}, 7(10):1030, 2017.

\bibitem{mourchid2021automatic}
Y.~Mourchid, M.~Donias, and Y.~Berthoumieu.
\newblock Automatic image colorization based on multi-discriminators generative adversarial networks.
\newblock In {\em 2020 28th European signal processing conference (EUSIPCO)}, pages 1532--1536. IEEE, 2021.

\bibitem{mourchid2016image}
Y.~Mourchid, M.~El~Hassouni, and H.~Cherif.
\newblock Image segmentation based on community detection approach.
\newblock {\em International Journal of Computer Information Systems and Industrial Management Applications}, 8:10--10, 2016.

\bibitem{mourchid2019movienet}
Y.~Mourchid, B.~Renoust, O.~Roupin, L.~V{\u{a}}n, H.~Cherifi, and M.~E. Hassouni.
\newblock Movienet: a movie multilayer network model using visual and textual semantic cues.
\newblock {\em Applied Network Science}, 4:1--37, 2019.

\bibitem{mourchid2023mr}
Y.~Mourchid and R.~Slama.
\newblock Mr-stgn: Multi-residual spatio temporal graph network using attention fusion for patient action assessment.
\newblock In {\em 2023 IEEE 25th International Workshop on Multimedia Signal Processing (MMSP)}, pages 1--6. IEEE, 2023.

\bibitem{qi2017pointnet}
C.~R. Qi, H.~Su, K.~Mo, and L.~J. Guibas.
\newblock Pointnet: Deep learning on point sets for 3d classification and segmentation.
\newblock In {\em Proceedings of the IEEE conference on computer vision and pattern recognition}, pages 652--660, 2017.

\bibitem{qi2017pointnet++}
C.~R. Qi, L.~Yi, H.~Su, and L.~J. Guibas.
\newblock Pointnet++: Deep hierarchical feature learning on point sets in a metric space.
\newblock {\em Advances in neural information processing systems}, 30, 2017.

\bibitem{rusu2009fast}
R.~B. Rusu, N.~Blodow, and M.~Beetz.
\newblock Fast point feature histograms (fpfh) for 3d registration.
\newblock In {\em 2009 IEEE international conference on robotics and automation}, pages 3212--3217. IEEE, 2009.

\bibitem{sain1996nature}
S.~R. Sain.
\newblock The nature of statistical learning theory, 1996.

\bibitem{schoenberg2010segmentation}
J.~R. Schoenberg, A.~Nathan, and M.~Campbell.
\newblock Segmentation of dense range information in complex urban scenes.
\newblock In {\em 2010 IEEE/RSJ International Conference on Intelligent Robots and Systems}, pages 2033--2038. IEEE, 2010.

\bibitem{tchapmi2017segcloud}
L.~Tchapmi, C.~Choy, I.~Armeni, J.~Gwak, and S.~Savarese.
\newblock Segcloud: Semantic segmentation of 3d point clouds.
\newblock In {\em 2017 international conference on 3D vision (3DV)}, pages 537--547. IEEE, 2017.

\bibitem{volk2014building}
R.~Volk, J.~Stengel, and F.~Schultmann.
\newblock Building information modeling (bim) for existing buildings—literature review and future needs.
\newblock {\em Automation in construction}, 38:109--127, 2014.

\bibitem{yuan20223d}
Q.~Yuan, Y.~Luo, and H.~Wang.
\newblock 3d point cloud recognition of substation equipment based on plane detection.
\newblock {\em Results in Engineering}, 15:100545, 2022.

\bibitem{zhang2020pointfilter}
D.~Zhang, X.~Lu, H.~Qin, and Y.~He.
\newblock Pointfilter: Point cloud filtering via encoder-decoder modeling.
\newblock {\em IEEE Transactions on Visualization and Computer Graphics}, 27(3):2015--2027, 2020.

\end{thebibliography}
\end{document}